\documentclass[11pt,a4paper]{article}
\usepackage[hyperref]{emnlp-ijcnlp-2019}
\usepackage{times}
\usepackage{latexsym}
\usepackage{multirow}
\usepackage{amsfonts}
\usepackage{booktabs}
\usepackage{hhline}
\usepackage{graphicx}

\usepackage{url}

\aclfinalcopy 

\newcommand{\safe}{{\sc safe}}
\newcommand{\unsafe}{{\sc offensive}}
\newcommand{\pone}{{\emph{[P1:]  \space}}}
\newcommand{\ptwo}{{\emph{[P2:]  \space}}}

\def\mylinesp{0.6em}

\title{Build it Break it Fix it for Dialogue Safety:\\ Robustness from Adversarial Human Attack}

\author{Emily Dinan \\
  Facebook AI Research \\
  {\small \tt edinan@fb.com} \\\And
  Samuel Humeau  \\
  Facebook AI Research \\
  {\small \tt samuel.humeau@gmail.com} \\\And
  Bharath Chintagunta \\
  Virginia Tech \\
  {\small \tt jaic4@vt.edu} \\
  \And
  Jason Weston \\
  Facebook AI Research \\
  {\small \tt jase@fb.com} \\}

\date{}

\begin{document}
\maketitle
\begin{abstract}
The detection of offensive language in the context of a dialogue has become an increasingly important application of natural language processing. The detection of trolls in public forums \cite{galan2016supervised}, and 
the deployment of chatbots in the public domain \cite{wolf2017we} are two examples that show the necessity of guarding against adversarially offensive behavior on the part of humans.
In this work, we develop a training scheme for a model to become robust to such human attacks by an iterative build it, break it, fix it strategy with humans and models in the loop. In detailed experiments we show this approach 
is considerably more robust than previous systems.
Further, we show that offensive language used within a conversation critically depends on the dialogue context, and cannot be viewed as a single sentence offensive detection task as in most previous work.
Our newly collected tasks and methods will be made open source and publicly available.

\end{abstract}

\section{Introduction}

The detection of offensive language has become an important topic
as the online community has grown, as so too have the number of bad actors 
\cite{cheng2017anyone}. Such behavior includes, but is not limited to, 
trolling in public discussion forums \cite{herring2002searching} and via social media \cite{silva2016analyzing,davidson2017automated}, 
employing hate speech that expresses prejudice against a particular group, 
or offensive language specifically targeting an individual.
Such actions can be motivated to cause harm from which the bad actor derives
enjoyment, despite negative consequences to others  \cite{bishop2014representations}.
As such, some bad actors
go to great lengths to both avoid detection and to achieve their goals
\cite{shachaf2010beyond}. In that context, any attempt to automatically
detect this behavior 
can be expected to be adversarially attacked by looking for weaknesses 
in the detection system,  which currently can easily be exploited as shown in 
\cite{hosseini2017deceiving, grondahl2018all}.
A further example, relevant to the natural langauge processing community, 
is the exploitation of weaknesses in machine learning models that {\em generate} 
text, to force them to emit offensive language. Adversarial attacks on the
Tay chatbot led to the developers shutting down the system \cite{wolf2017we}.

In this work, we study the detection of offensive language in 
dialogue with models that are robust to adversarial attack.
We develop an automatic approach to the ``Build it Break it Fix it'' strategy 
originally adopted for writing secure programs \cite{ruef2016build}, and  the ``Build it Break it'' approach consequently adapting it for NLP \cite{ettinger2017towards}.
In the latter work, two teams of researchers, ``builders'' and ``breakers'' were used
to first create sentiment and semantic role-labeling systems and then construct examples that find their faults. In this work we instead 
fully automate such an approach 
using crowdworkers as the humans-in-the-loop, 
and also apply a fixing stage where models are retrained to improve them. Finally, 
we repeat the whole build, break, and fix sequence over a number of iterations.

We show that such an approach provides more and more robust systems over the fixing iterations. 
Analysis of the type of data collected in the iterations of the break it phase shows clear distribution changes, moving away from simple use of profanity and other obvious offensive words to utterances
that require understanding of world knowledge, figurative 
language, and use of negation to detect if they are offensive or not. 
Further, data collected in the context of a dialogue rather than a sentence without context provides more sophisticated attacks. We show that model architectures
that use the dialogue context efficiently perform much better than systems that do not, where the latter has been the main focus of existing research
\cite{personal_attack,davidson2017automated,zampieri2019semeval}.

Code for our entire build it, break it, fix it algorithm will be made open source,
complete with model training code and crowdsourcing interface for humans. 
Our data and trained models will also be made available for the community.

\section{Related Work}

The task of detecting offensive language has been studied across a variety of content classes.  Perhaps the most commonly studied class is hate speech, but work has also covered bullying, aggression, and toxic comments \cite{zampieri2019semeval}. 

To this end, various datasets have been created to benchmark progress in the field. In hate speech detection, recently \citet{davidson2017automated} compiled and released a dataset of over 24,000 tweets labeled as containing hate speech, offensive language, or neither. The TRAC shared task on Aggression Identification, a dataset of over 15,000 Facebook comments labeled with varying levels of aggression, was released as part of a competition \cite{kumar-etal-2018-benchmarking}. In order to benchmark toxic comment detection, The Wikipedia Toxic Comments dataset (which we study in this work) was collected and extracted from Wikipedia Talk pages and featured in a Kaggle competition \cite{personal_attack, ToxicComment}. Each of these benchmarks examine only single-turn utterances, outside of the context in which the language appeared. In this work we recommend that future systems should move beyond classification of singular utterances and use contextual information to help identify offensive language.

Many approaches have been taken to solve these tasks 
-- from linear regression and SVMs to deep learning \cite{noever2018machine}. The best performing systems in each of the competitions mentioned above (for aggression and toxic comment classification) used deep learning approaches such as LSTMs and CNNs \cite{kumar-etal-2018-benchmarking,ToxicComment}. In this work we consider a large-pretrained transformer model which has been shown to perform well on many downstream NLP tasks \cite{Devlin2018BERTPO}.

The broad class of adversarial training is currently a hot topic in 
machine learning \cite{goodfellow2014generative}. Use cases include training image generators 
\cite{brock2018large} as well as image classifiers to be robust to adversarial examples \cite{liu2019perceptual}.
These methods find the breaking examples algorithmically, rather than by using humans breakers
as we do.
Applying the same approaches to NLP tends to be more challenging because, unlike for images, even small changes to a sentence can cause a large change in the meaning of that sentence, which a human can detect but a lower quality model cannot. Nevertheless algorithmic
approaches have been attempted, for example in 
 text classification \cite{ebrahimi2017hotflip}, machine translation \cite{belinkov2017synthetic}, dialogue generation tasks \cite{li2017adversarial} and reading comprehension \cite{jia2017adversarial}. The latter was particularly effective at proposing a more difficult version of the popular SQuAD dataset.

As mentioned in the introduction, our approach takes inspiration from ``Build it Break it'' approaches which have been successfully tried in other domains \cite{ruef2016build,ettinger2017towards}. Those approaches advocate finding faults in systems by having humans look for insecurities (in software) or prediction failures (in models), but 
do not advocate an automated approach as we do here.
Our work is also closely connected to the ``Mechanical Turker Descent'' algorithm detailed
in \cite{Yang2018MasteringTD} where language to action pairs were collected from crowdworkers 
by incentivizing them with a game-with-a-purpose technique: a crowdworker receives a bonus if their contribution results in better models than another crowdworker. We did not gamify our approach in
this way, but still our approach has commonalities in the round-based improvement of models through
crowdworker interaction.

\section{Baselines: Wikipedia Toxic Comments}
\label{section:baselines}

In this section we describe the publicly available data that we have used to bootstrap our {\em build it break it fix it} approach. We also compare our model choices with existing work and clarify the metrics chosen to report our results.

\label{section:baseline}

\paragraph{Wikipedia Toxic Comments} The Wikipedia Toxic Comments dataset (WTC) has been collected in a common effort from the Wikimedia Foundation and Jigsaw \cite{personal_attack} to identify personal attacks online. The data has been extracted from the Wikipedia Talk pages, discussion pages where editors can discuss improvements to articles or other Wikipedia pages. We considered the version of the dataset that corresponds to the Kaggle competition: ``Toxic Comment Classification Challenge" \cite{ToxicComment} which features 7 classes of toxicity: toxic, severe toxic, obscene, threat, insult, identity hate and non-toxic. In the same way as in \cite{khatri_offensive}, every label except non-toxic is grouped into a class \unsafe\ while the non-toxic class is kept as the \safe\ class. In order to compare our results to \cite{khatri_offensive}, we similarly split this dataset to dedicate 10\% as a test set. 80\% are dedicated to train set while the remaining 10\% is used for validation. Statistics on the dataset are shown in Table \ref{table:WTC_stats}.

\paragraph{Models} We establish baselines using two models. The first one is a binary classifier built on top of a large pre-trained transformer model. We use the same architecture as in BERT \cite{Devlin2018BERTPO}.
We add a linear layer to the output of the first token ([CLS]) to produce a final binary classification.
We initialize the model using the weights provided by \cite{Devlin2018BERTPO} corresponding to ``BERT-base". The transformer is composed of 12 layers with hidden size of 768 and 12 attention heads. We fine-tune the whole network on the classification task.
We also compare it the fastText classifier \cite{joulin2017bag} for which a given sentence is encoded as the average of individual word vectors that are pre-trained on a large corpus issued from Wikipedia. A linear layer is then applied on top to yield a binary classification.

\paragraph{Experiments}
We compare the two aforementioned models with \cite{khatri_offensive} who conducted their experiments with a BiLSTM with GloVe pre-trained word vectors \cite{glove}. Results are listed in Table \ref{table:baseline_results} and we compare them using the weighted-F1, i.e. the sum of F1 score of each class weighted by their frequency in the dataset. We also report the F1 of the \unsafe-class which is the metric we favor within this work, although we report both. (Note that throughout the paper, the notation F1 is always referring to \unsafe-class F1.) Indeed, in the case of an imbalanced dataset such as Wikipedia Toxic Comments where most samples are \safe, the weighted-F1 is closer to the F1 score of the \safe\ class while we focus on detecting \unsafe\ content. Our BERT-based model outperforms the method from \citet{khatri_offensive}; throughout the rest of the paper, we use the BERT-based architecture in our experiments. In particular, we used this baseline trained on WTC to bootstrap our 
approach, to be described subsequently.

\begin{table}[t]
\center
\small
\begin{tabular}{lccc}
\toprule
 & Train & Valid & Test\\
\midrule
\safe & 89.8\% & 89.7\% & 90.1\%\\
\unsafe & 10.2\% & 10.3\% & 9.1\%\\
Total & 114656 & 15958 & 15957\\
\bottomrule
\end{tabular}
\caption{Dataset statistics for our splits of Wikipedia Toxic Comments.}
\label{table:WTC_stats}
\end{table}

\begin{table}[t]
\center
\small
\begin{tabular}{lcc}
\toprule
 & \unsafe\ F1 & Weighted F1\\
\midrule
fastText & 71.4\% & 94.8\% \\
BERT-based & 83.4\% & 96.7\% \\
\cite{khatri_offensive} & -  & 95.4\% \\
\bottomrule
\end{tabular}
\caption{Comparison between our models based on fastText and BERT with the BiLSTM used by \cite{khatri_offensive} on Wikipedia Toxic Comments.}
\label{table:baseline_results}
\end{table}

\begin{figure}[h]
    \centering
    \includegraphics[width=0.47\textwidth]{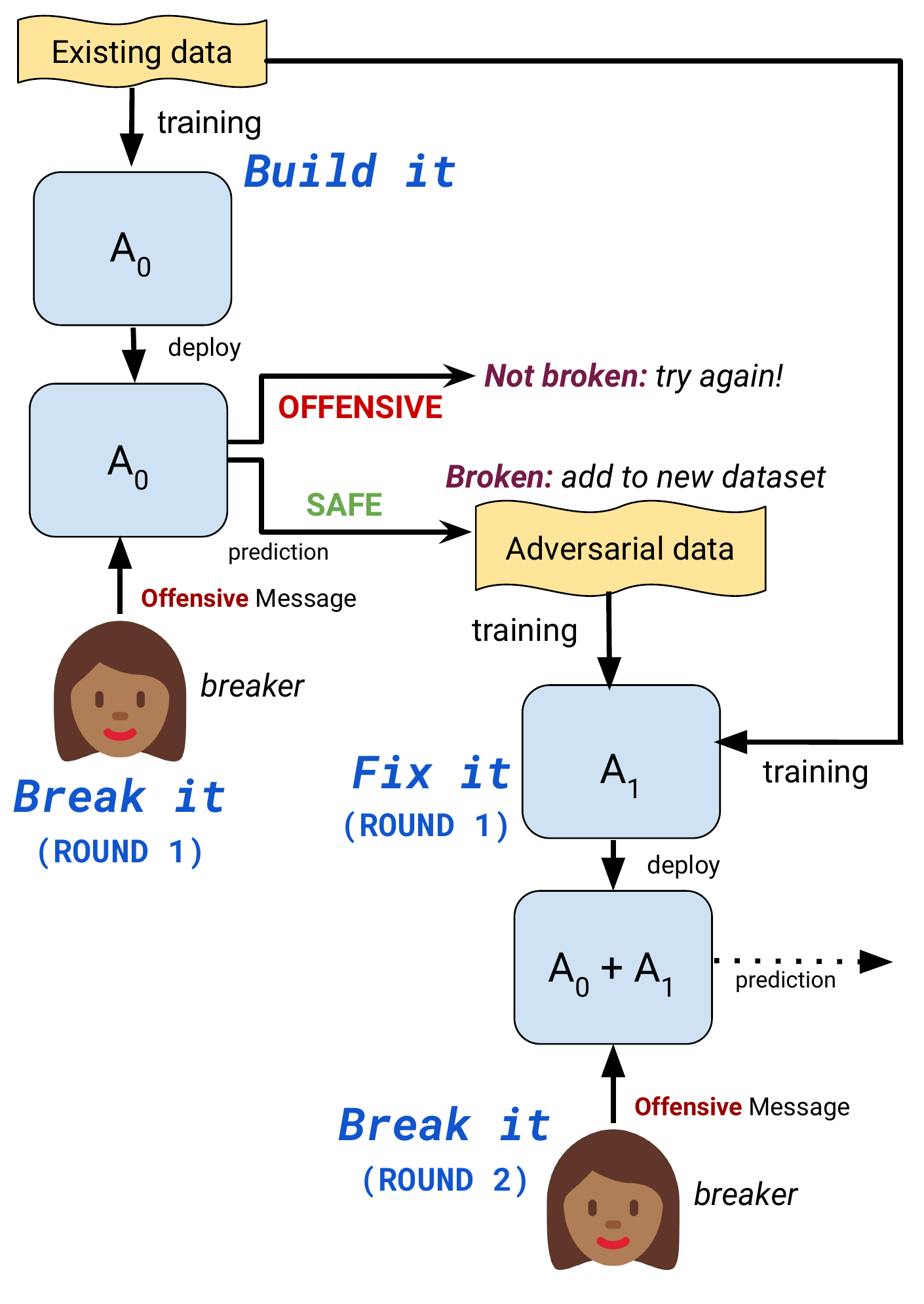}
    \caption{The build it, break it, fix it algorithm we use to iteratively train better models $A_0, \dots, A_N$. In experiments we perform $N=3$ iterations of the break it, fix it loop for the single-turn utterance detection task, and a further iteration for the multi-turn task in a dialogue context setting. }
    \label{fig:diagram}
\end{figure}

\section {Build it Break it Fix it Method}

In order to train models that are robust to adversarial behavior, we posit that it is crucial collect and train on data that was collected in an adversarial manner. We propose the following
automated build it, break it, fix it algorithm: 

\begin{enumerate}
    \item {\bf Build it:} Build a model capable of detecting \unsafe\ messages.  This is our best-performing BERT-based model trained on the Wikipedia Toxic Comments dataset described in the previous section. We refer to this model throughout as $A_0$. 
    \item {\bf Break it:} Ask crowdworkers to try to ``beat the system" by submitting messages that our system ($A_0$) marks as \safe\ but that the worker considers to be \unsafe.
    \item {\bf Fix it:} Train a new model on these collected examples in order to be  more robust to these adversarial attacks.
    \item {\bf Repeat:} Repeat, deploying the newly trained model in the {\bf break it} phase, then \textbf{fix it} again.
\end{enumerate}

See Figure \ref{fig:diagram} for a visualization of this process. 

\subsection{Break it Details}

\paragraph{Definition of \unsafe } Throughout data collection, we characterize \unsafe\ messages for users as messages that would not be ``ok to send in a friendly conversation with someone you just met online." 
We use this specific language in an attempt to capture various classes of content that would be considered unacceptable in a friendly conversation, without imposing our own definitions of what that means.  
The phrase ``with someone you just met online" was meant to mimic the setting of a public forum.

\paragraph{Crowderworker Task} We ask crowdworkers to try to ``beat the system" by submitting messages that our system marks as \safe\ but that the worker considers to be \unsafe. For a given round, workers earn a ``game'' point each time they are able to ``beat the system," or in other words, trick the model by submitting \unsafe\ messages that the model marks as \safe. Workers earn up to 5 points each round, and have two tries for each point: we allow multiple attempts per point so that workers can get feedback from the models and better understand their weaknesses. The points serve to indicate 
success to the crowdworker and motivate to achieve high scores, but have no other meaning  (e.g. no monetary value as in \cite{Yang2018MasteringTD}).
More details regarding the user interface and instructions can be found in Appendix \ref{appendix:datacollection}. 

\paragraph{Models to Break}
During round $1$, workers try to break the baseline model $A_0$, trained on Wikipedia Toxic Comments. For rounds $i$, $i > 1$, workers must break both the baseline model and the model from the previous ``fix it" round, which we refer to as $A_{i-1}$. In that case, the worker must submit messages that both $A_0$ and $A_{i-1}$ mark as \safe\ but which the worker considers to be \unsafe.

\subsection{Fix it Details}

During the ``fix it" round, we update the models with the newly collected adversarial data from the ``break it" round.

The training data consists of all previous rounds of data, so that model $A_i$ is trained on all rounds $n$ for $n \leq i$, as well as the Wikipedia Toxic Comments data. We split each round of data into train, validation, and test partitions. The validation set is used for hyperparameter selection. The test sets are used to measure how robust we are to new adversarial attacks. With increasing round $i$, $A_i$ should become more robust to increasingly complex human adversarial attacks.

\begin{table*}[t!]
\small
\begin{center}
\textbf{Single-Turn Adversarial (Round $1$) and Standard Task \unsafe\ Examples }   \\
\begin{tabular}{lccccccc}
\toprule
& contains & non-profane & contains & contains & requires & contains\\
&  profanity & offending words &  negation & figurative language &  world knowledge &  sarcasm \\
\midrule
Standard    & 13\% & 12\%  &  12\% & 11\% & 8\%  & 3\%   \\
Adversarial & 0\% & 5\% &  23\% & 19\% & 14\% & 6\%   \\
\end{tabular}
\end{center}
\caption{Language analysis of the single-turn  \emph{standard} and \emph{adversarial} (round 1) tasks by human annotation of various language properties. Standard collection examples contain more words found in an offensive words list, while \emph{adversarial} examples require more sophisticated  language understanding.
\label{table:singleexamples}
}
\end{table*}

\section{Single-Turn Task}

  We first consider a single-turn set-up, i.e. detection of offensive language in one utterance, with no  dialogue context or conversational history. 

\subsection{Data Collection}

\paragraph{Adversarial Collection}

We collected three rounds of data with the build it, break it, fix it algorithm described in the previous section. Each round of data consisted of 1000 examples, leading to 3000 single-turn adversarial examples in total. For the remainder of the paper, we refer to this method of data collection as the \emph{adversarial method}.

\paragraph{Standard Collection}
In addition to the \emph{adversarial method}, we also collected data in a non-adversarial manner in order to directly compare the two set-ups. In this method -- which we refer to as the \emph{standard method}, we simply ask crowdworkers to submit messages that they consider to be \unsafe. There is no model to break. Instructions are otherwise the same.

In this set-up, there is no real notion of ``rounds", but for the sake of comparison we refer to each subsequent 1000 examples collected in this manner as a ``round". We collect 3000 examples -- or three rounds of data. We refer to a model trained on rounds $n \leq i$ of the \emph{standard} data as $S_i$.

\subsubsection{Task Formulation Details}
\label{section:taskimpl}

Since all of the collected examples are labeled as \unsafe, to make this task a binary classification problem, we will also add  \safe\ examples to it.

The ``safe data" is comprised of utterances from the ConvAI2 chit-chat task \cite{convai2comp,personachat}  
which consists of pairs of humans getting to know each other by discussing their interests.
Each utterance we used was reviewed by two independent crowdworkers and labeled as \safe,
with the same characterization of \safe\ 
as described before.

For each partition (train, validation, test), the final task has a ratio of 9:1 \safe\ to \unsafe\ examples, mimicking the division of the Wikipedia Toxic Comments dataset used for training our baseline models. Dataset statistics for the final task can be found in Table \ref{table:datastats}. We refer to these tasks -- with both \safe\ and \unsafe\ examples -- as the \emph{adversarial} and \emph{standard} tasks.

\begin{table}[h]
\small
\begin{center}
\begin{tabular}{lllll}
\toprule
 & \% with & \% with & avg. \# & avg. \# \\
 & profanity & ``not" & chars & tokens \\ 
\midrule
{\bf Std. (Rnds $1$-$3$)} & 18.2 & 2.8 & 48.6 & 9.4 \\
{\bf Adv. Rnd $1$} & 2.6 & 5.8 & 53.7 & 10.3 \\
{\bf Adv. Rnd $2$} & 1.5 & 5.5 & 44.5 & 9 \\
{\bf Adv. Rnd $3$} & 1.2 & 9.8 & 45.7 & 9.3 \\
{\bf Multi-turn Adv.} & 1.6 & 4.9 & 36.6 & 7.8 \\ 
\bottomrule
\end{tabular}
\end{center}
\caption{Percent of \unsafe\ examples in each task containing profanity, the token ``not", as well as the average number of characters and tokens in each example. Rows 1-4 are the single-turn task, and the last row is the multi-turn task.
Later rounds have less profanity and more use of negation as human breakers have to find more sophisticated language to adversarially attack our models.
}
\label{table:datadist}
\label{tablestats}
\small
\begin{center}
\begin{tabular}{lrrrr}
\toprule
  Rounds \{1, 2 and 3\}  & Train & Valid & Test \\
\midrule
{\bf \safe\ Examples} & 21,600 & 2700 & 2700  \\
{\bf \unsafe\ Examples} & 2400 & 300 & 300  \\ 
{\bf Total Examples} & 24,000 & 3,000 & 3,000  \\
\bottomrule
\end{tabular}
\end{center}
\caption{Dataset statistics for the single-turn rounds of the \emph{adversarial} task data collection.
There are three rounds in total  all of identical size, 
hence the numbers above can be divided for individual statistics.
The \emph{standard} task is  an additional dataset of exactly the same size as above.} 
\label{table:datastats}
\end{table}

\begin{table*}[t!]
\small
\center
\setlength\tabcolsep{9pt} 
{\renewcommand{\arraystretch}{1.2}
\begin{tabular}{llrrrrrrr}
\toprule
  & & {WTC Baseline} & \multicolumn{3}{c}{ \emph{Standard} models} & \multicolumn{3}{c}{\emph{Adversarial} models} \\
 \cmidrule(lr){3-3}\cmidrule(lr){4-6} \cmidrule(lr){7-9}
{Task Type} & {Task Round} & $A_0$  & $S_1$ & $S_2$ & $S_3$ & $A_1$ & $A_2$ & $A_3$ \\
\midrule
WTC             & - & {\bf 83.3} & 80.6 & 81.1 & 82.1 & 81.3 & 78.9 & 78.0 \\
\midrule

                {\emph{Standard Task}} & All (1-3) & 68.1 & 83.3 & 85.8 & {\bf 88.0} & 83.0 & 85.3 & 83.7 \\ 
\midrule
\multirow{4}{*}{{\emph{Adversarial Task}}}                & 1 &0.0& 51.7 & 69.3 & 68.6 & 71.8 & {\bf 79.0} & 78.2  \\ 
 & 2 &0.0 & 10.8 & 26.4 & 31.8 &0.0& {\bf 64.4} & 62.1 \\ 
               & 3 &0.0& 12.3 & 17.1 & 13.7 & 32.1 &0.0& {\bf 59.9} \\ 
                   & All (1-3) &0.0& 27.4 & 41.7 & 41.8 & 40.6 & 55.5 & {\bf 67.6}  \\
\bottomrule
\end{tabular}
\caption{Test performance of best standard models trained on \emph{standard} task rounds (models $S_i$ for each round $i$) and best \emph{adversarial} models trained on adversarial task rounds (models $A_i$). All models are evaluated using \unsafe-class F1 on each round of both the \emph{standard} task and \emph{adversarial} task. $A_0$ is the baseline model trained on the existing Wiki Toxic Comments (WTC) dataset. Adversarial models prove to be more robust than standard ones against attack (Adversarial Task 1-3), 
while still performing reasonably on Standard and WTC tasks.
}
\label{table:sensitive_f1_results}
}
\end{table*}

\subsubsection{Model Training Details}
\label{section:taskimpl2}

Using the BERT-based model architecture described in Section \ref{section:baseline}, we trained models on each round of the \emph{standard} and \emph{adversarial} tasks, multi-tasking with the Wikipedia Toxic Comments task.
We weight the multi-tasking with a mixing parameter which is also tuned on the validation set. 
Finally, after training weights with the cross entropy loss, 
we adjust the final bias also using the validation set.
We optimize for the sensitive class (i.e.~\unsafe-class) F1 metric on the \emph{standard} and \emph{adversarial} validation sets respectively.

For each task (\emph{standard} and \emph{adversarial}), on round $i$, we train on data from all rounds $n$ for $n \leq i$ and optimize for performance on the validation sets $n \leq i$.

\subsection{Experimental Results}

We conduct experiments comparing the \emph{adversarial} and \emph{standard} methods. We break down the results into ``break it" results comparing the data collected and ``fix it" results comparing the models obtained.

\subsubsection{Break it Phase}

Examples obtained from both the \emph{adversarial} and \emph{standard} collection methods were found to be clearly offensive, but we note several differences in the distribution of examples from each task, shown in Table \ref{table:datadist}. First, examples from the \emph{standard} task tend to contain more profanity. Using a list of common English obscenities and otherwise bad words\footnote{https://github.com/LDNOOBW/List-of-Dirty-Naughty-Obscene-and-Otherwise-Bad-Words}, in Table \ref{table:datadist} we calculate the percentage of examples in each task containing such obscenities, and see that the \emph{standard} examples contain at least seven times as many as each round of the \emph{adversarial} task. Additionally, in previous works, authors have observed that classifiers struggle with negations \cite{hosseini2017deceiving}. This is borne out by our data: examples from the single-turn \emph{adversarial} task more often contain the token ``not" than examples from the \emph{standard} task, indicating that users are easily able to fool the classifier with negations. 

We also anecdotally see figurative language such as ``snakes hiding in the grass'' in the adversarial data, which contain no individually offensive words, the offensive nature is captured  by reading the entire sentence. 
Other examples require sophisticated world knowledge such as that many cultures consider eating cats to be offensive.
To quantify these differences, we performed a blind human 
annotation of a sample of the data, 100 examples of standard and 100 examples of
adversarial round 1. Results are shown in Table \ref{table:singleexamples}.
Adversarial data was indeed found to contain less profanity, fewer non-profane but offending words (such as ``idiot''), more figurative language, and to require more world knowledge.

We note that, as anticipated, the task becomes more challenging for the crowdworkers with each round, indicated by the decreasing average scores in Table \ref{table:turkstats}. In round $1$, workers are able to get past $A_0$ most of the time -- earning an average score of $4.56$ out of $5$ points per round -- showcasing how susceptible this baseline is to adversarial attack despite its relatively strong performance on the Wikipedia Toxic Comments task. By round $3$, however, workers struggle to trick the system, earning an average score of only $1.6$ out of $5$. A finer-grained assessment of the worker scores can be found in Table \ref{table:turkstatsfull} in the appendix.

\begin{table}[t]

\begin{center}

\footnotesize
\begin{tabular}{lrrrr}
 \toprule
& \multicolumn{3}{c}{Single-Turn} & {Multi} \\
\cmidrule(lr){2-4}\cmidrule(lr){5-5}
Round & \multicolumn{1}{c}{1} &  \multicolumn{1}{c}{2} &  \multicolumn{1}{c}{3} & \multicolumn{1}{c}{(``4")}\\
\midrule
Avg. score (0-5)  & 4.56 & 2.56 & 1.6 & 2.89\\ 
\bottomrule
\end{tabular}
\end{center}
\caption{Adversarial data collection worker scores. Workers received a score out of 5 indicating how often (out of 5 rounds) they were able to get past our classifiers within two tries. In later single-turn rounds it is harder to defeat our models, but switching to multi-turn makes this easier again as new attacks can be found by using the dialogue context.}
\label{table:turkstats}
\end{table}

\subsubsection{Fix it Phase}

Results comparing the performance of models trained on the \emph{adversarial} ($A_i$) and \emph{standard} ($S_i$)    tasks are summarized in Table \ref{table:sensitive_f1_results}, with further results in Table \ref{table:fulltable} in Appendix \ref{appendix:experiments}. 
The adversarially trained models $A_i$ prove to be more robust to adversarial attack: on each round of adversarial testing they outperform standard models $S_i$.

Further, note that the \emph{adversarial} task becomes harder with each subsequent round. In particular, the performance of the standard models $S_i$ rapidly deteriorates between round $1$ and round $2$ of the \emph{adversarial} task.
This is a clear indication that models need to train on adversarially-collected data to be robust to adversarial behavior.

\emph{Standard} models ($S_i$), trained on the \emph{standard} data, tend to perform similarly to the \emph{adversarial} models ($A_i$) as measured on the \emph{standard} test sets, with the exception of training round $3$, in which 
$A_3$ fails to improve on this task, likely due to being too optimized for adversarial tasks. The \emph{standard} models $S_i$, on the other hand, are improving with subsequent rounds as they have more training data of the same distribution as the evaluation set.
Similarly, our baseline model performs best on its own test set, but other models are not far behind.

\begin{table*}[t]
\small
\begin{center}
\begin{tabular}{ll}
\toprule
\multicolumn{2}{l}{ \bf Multi-Turn Adversarial Task Examples} \\
\midrule
{\bf Context} & \pone greetings and salutations my friend   \\
& \ptwo hi how are you doing? \\ 
& \pone i'm well. enjoying some r r from the navy station \\
& \ptwo i love to cook. do you cook? \\ 
& \pone i do on occasion. i love to read more though. \\
{\bf \unsafe\ Response } & \ptwo {\bf reading the menu doesnt count}    \\ 

\midrule
{\bf Context} & \pone hey how are you   \\
& \ptwo good but i keep thinking about the time i almost played for nba \\ 
& \pone aww why did not you play \\
& \ptwo lost my nerve i guess. what do you do? \\ 
& \pone oh that's to bad. i work at a gas station \\
{\bf \unsafe\ Response} & \ptwo {\bf i need to follow my dream to not work in a gas station}   \\ 
\bottomrule
\end{tabular}
\end{center}
\caption{Examples from the \emph{multi-turn adversarial} task. Responses can be offensive only in context.}
\label{table:multiexamples}
\end{table*}

\begin{table}[t]
\small
\begin{center}
\begin{tabular}{lrrrr}
\toprule
  \bf{Multi-Turn Adversarial Task}  & Train & Valid & Test \\
\midrule
{\bf \safe\ Examples} & 21,600 & 2,700 & 2,700  \\
{\bf \unsafe\ Examples} & 2,400 & 300 & 300  \\ 
{\bf Total Examples} & 24,000 & 3,000 & 3,000  \\
\bottomrule
\end{tabular}
\end{center}
\caption{
\emph{Multi-turn adversarial} task data statistics.
}
\label{table:multidatastats}
\end{table}

Finally, we remark that all scores of $0$ in Table \ref{table:sensitive_f1_results} are by design, as for round $i$ of the \emph{adversarial} task, both $A_0$ and $A_{i-1}$ classified each example as 
\safe\ during the `break it' data collection phase.

\section{Multi-Turn Task}

In most real-world applications, we find that adversarial behavior occurs in context -- whether it is in the context of a one-on-one conversation, a comment thread, or even an image. In this work we focus on
offensive utterances within the context of two-person dialogues.
For dialogue safety we posit it is important to move beyond classifying single utterances, as it may be the case that an utterance is entirely innocuous on its own but extremely offensive in the context of the previous dialogue history. For instance, ``Yes, you should definitely do it!" is a rather inoffensive message by itself, but most would agree that it is a hurtful response to the question ``Should I hurt myself?"

\begin{table}[h!]
\begin{center}
\small
\setlength\tabcolsep{8pt} 
\begin{tabular}{lcc}
\toprule
\multicolumn{3}{c}{\bf{Multi-Turn Adversarial Task Results}} \\
\midrule
  & F1 & Weighted-F1 \\
\midrule
{\bf fastText} \\    
~with context & 23.6 $\pm$ 1.9 & 85.9 $\pm$ 0.5  \\ 
~without context & 37.1 $\pm$ 2.6 & 88.8 $\pm$ 0.6 \\ 
\midrule
{\bf BERT-based} \\
{\em (no segments)} \\    
~with context & 60.5 $\pm$ 1.3  & 92.2 $\pm$ 0.3 \\ 
~without context & 56.8 $\pm$ 1.6 & 90.6 $\pm$ 0.7 \\ 
\midrule
{\bf BERT-based} \\
{\em (dialogue segments)} \\    
~with context &  66.4 $\pm$ 2.2 & 93.2 $\pm$ 0.4 \\ 
~without context & 59.0 $\pm$ 2.5 & 91.2 $\pm$ 0.8 \\ 
\bottomrule
\end{tabular}
\end{center}
\caption{Results of experiments on the multi-turn  \emph{adversarial} task. We denote the average and one standard deviation from the results of five runs. 
Models that use the context as input (``with context'') perform better. Encoding this in the architecture as well (via BERT {\em dialogue segment} features) gives us the best results.
}
\label{table:multiturnresults}
\end{table}

\subsection{Task Implementation}

To this end, we collect data by asking crowdworkers to try to ``beat" our best single-turn classifier (using the model that performed best on rounds 1-3 of the \emph{adversarial} task, i.e., $A_3$), in addition to our baseline classifier $A_0$. The workers are shown truncated pieces of a conversation from the ConvAI2 chit-chat task, and asked to continue the conversation with \unsafe\ responses that our classifier marks as \safe. As before, workers have two attempts per conversation to try to get past the classifier and are shown five conversations per round. They are given a score (out of five) at the end of each round indicating the number of times they successfully fooled the classifier.

We collected 3000 offensive examples in this manner. As in the single-turn set up, we  combine this data with \safe\ examples with a ratio of 9:1  \safe\ to \unsafe~for classifier training. The safe examples are dialogue examples from  ConvAI2  for which the responses were reviewed by two independent crowdworkers and labeled as \safe, as in the s single-turn task set-up. We refer to this overall task as the \emph{multi-turn adversarial} task. Dataset statistics are given in Table \ref{table:multidatastats}.

\subsection{Models}

To measure the impact of the context, we train models on this dataset with and without the given context. We use the fastText and the BERT-based model described in Section \ref{section:baseline}. In addition, we build a BERT-based model variant that splits the last utterance (to be classified) and the rest of the history into two dialogue segments. Each segment is assigned an embedding and the input provided to the transformer is the sum of word embedding and segment embedding, replicating the setup of the Next Sentence Prediction that is used in the training of BERT \cite{Devlin2018BERTPO}.

\subsection{Experimental Results}

\subsubsection{Break it Phase}

During data collection, we observed that workers had an easier time bypassing the classifiers than in the single-turn set-up. See Table \ref{table:turkstats}. In the single-turn set-up, the task at hand gets harder with each round -- the average score of the crowdworkers decreases from $4.56$ in round $1$ to $1.6$ in round $3$. Despite the fact that we are using our best single-turn classifier in the multi-turn set-up ($A_3$), the task becomes easier: the average score per round is $2.89$. This is because the workers are often able to use contextual information to suggest something offensive rather than say something offensive outright. See examples of submitted messages in Table \ref{table:multiexamples}. Having context also allows one to express something offensive more efficiently: the messages supplied by workers in the multi-turn setting were significantly shorter on average, see  Table \ref{table:datadist}.

\subsubsection{Fix it Phase}

During training, we multi-tasked the \emph{multi-turn adversarial} task with the Wikipedia Toxic Comments task as well as the single-turn \emph{adversarial} and \emph{standard} tasks. We average the results of our best models from five different training runs. The results of these experiments are given in Table \ref{table:multiturnresults}. 

As we observed during the training of our baselines in Section \ref{section:baselines}, the fastText model architecture is ill-equipped for this task relative to our BERT-based architectures. The fastText model performs worse given the dialogue context (an average of 23.56 \unsafe-class F1 relative to 37.1) than without, likely because its bag-of-embeddings representation is too simple to take the context into account. 

We see the opposite with our BERT-based models, indicating that more complex models are able to effectively use the contextual information to detect whether the response is \safe\ or \unsafe. With the simple BERT-based architecture (that does not split the context and the utterance into separate segments), we observe an average of a 3.7 point increase in \unsafe-class F1 with the addition of context. When we use segments to separate the context from the utterance we are trying to classify, we observe an average of a 7.4 point increase in \unsafe-class F1. Thus, it appears that the use of contextual information to identify \unsafe\ language is critical to making these systems robust, and improving the model architecture to take account of this has large impact.

\section{Conclusion}

We have presented an approach to build more robust 
offensive language detection systems in the context of 
a dialogue. We proposed a build it, break it, fix it, and then repeat strategy, whereby humans attempt to break the models we built, and we use the broken examples to fix the models. 
We show this results in far more nuanced language than in
existing datasets. The adversarial data includes less profanity, which existing classifiers can pick up on, and is instead offensive due to figurative language, negation, and  by requiring more
world knowledge, which all make current classifiers fail. Similarly, offensive language in the context of a {\em dialogue} is also more nuanced than stand-alone offensive utterances.
We show that  classifiers that learn from these more complex
examples are indeed more robust to attack, and that using the dialogue context gives improved performance if the model 
architecture takes it into account.

In this work we considered a  
binary problem (offensive or safe). 
Future work could 
consider classes of offensive  language separately 
\cite{zampieri2019semeval}, or explore 
other dialogue tasks, e.g. 
 from social media or forums.
Another interesting direction is to explore how 
our build it, break it, fix it strategy
would similarly
apply to make neural generative models safe 
\cite{Henderson:2018:ECD:3278721.3278777}. 


\bibliography{emnlp-ijcnlp-2019}
\bibliographystyle{acl_natbib}

\newpage
~
~
\newpage

\appendix
\section{Additional Experimental Results}
\subsection{Additional Break It Phase Results}

Additional results regarding the crowdworkers' ability to ``beat" the classifiers are reported in Table \ref{table:turkstatsfull}. In particular, we report the percent of messages sent by the crowdsource workers that were marked \safe\ and \unsafe\ by both $A_0$ and $A_{i-1}$. We note that very infrequently ($<1\%$ of the time) a message was marked \unsafe\ by $A_0$ but \safe\ by $A_{i-1}$, showing that $A_0$ was relatively ineffective at catching adversarial behavior.

\begin{table}[h]
\begin{center}

\footnotesize
\begin{tabular}{lrrrr}
 \toprule
& \multicolumn{3}{c}{Single-Turn} & {Multi} \\
\cmidrule(lr){2-4}\cmidrule(lr){5-5}
Round & \multicolumn{1}{c}{1} &  \multicolumn{1}{c}{2} &  \multicolumn{1}{c}{3} & \multicolumn{1}{c}{(``4")}\\
\midrule
Avg. score (0-5)  & 4.56 & 2.56 & 1.6 & 2.89\\ 
\midrule
$A_0$: \unsafe\ and  & \multirow{2}{*}{-} & \multirow{2}{*}{0.6\%} & \multirow{2}{*}{0.8\%} & \multirow{2}{*}{1.4\%}\\ 
$A_{i-1}$: \safe & & & & \\ 
\addlinespace[\mylinesp]
$A_0$: \safe\ and & \multirow{2}{*}{-} & \multirow{2}{*}{44.7\%} & \multirow{2}{*}{64.9\%} & \multirow{2}{*}{17.7\%}\\
$A_{i-1}$: \unsafe & & & & \\ 
\addlinespace[\mylinesp]
$A_0$: \unsafe\ and & \multirow{2}{*}{25.7\%} & \multirow{2}{*}{23.7\%} & \multirow{2}{*}{16.1\%} & \multirow{2}{*}{4.1\%}\\
$A_{i-1}$: \unsafe & & & & \\ 
\addlinespace[\mylinesp]
$A_0$: \safe\ and & \multirow{2}{*}{74.3\%} & \multirow{2}{*}{31.1\%} & \multirow{2}{*}{18.3\%} & \multirow{2}{*}{76.8\%}\\
$A_{i-1}$: \safe  & & & & \\ 
\bottomrule
\end{tabular}
\end{center}
\caption{Adversarial data collection statistics. $A_0$ is the baseline model, trained on the Wikipedia Toxic Comments dataset. $A_{i-1}$ is the model for round $i$, trained on the adversarial data for rounds $n \leq i-1$. In the case of the multi-turn set-up, $A_{i-1}$ is $A_3$.}
\label{table:turkstatsfull}
\end{table}

In Table \ref{table:singleexamples_classstats}, we report the categorization of examples into classes of offensive language from the blind human annotation of round 1 of the single-turn \emph{adversarial} and \emph{standard} data. We observe that in the \emph{adversarial} set-up, there were fewer examples of bullying language but more examples targeting a protected class.

\begin{table*}[t!]
\small
\begin{center}
\textbf{Single-Turn Adversarial and Standard Task \unsafe\ Examples (Round $1$)}   \\
\begin{tabular}{lcccccc}
\toprule
 & protected & non-protected && &\\
 &  class &  class & bullying  &  sexual & violent \\ 
\midrule
Standard    & 16\% & 18\% & 60\% & 8\%  & 10\% \\
Adversarial & 25\% & 16\% & 28\% &  14\% & 15\% \\
\bottomrule
\end{tabular}
\end{center}
\caption{Human annotation of 100 examples from each the single-turn  \emph{standard} and \emph{adversarial} (round 1) tasks into offensive classes.
\label{table:singleexamples_classstats}
}
\end{table*}

\subsection{Additional Fix It Phase Results}

We report F1, precision, and recall for the \unsafe\ class, as well as weighted-F1 for models $S_i$ and $A_i$ on the single-turn \emph{standard} and \emph{adversarial} tasks in Table \ref{table:fulltable}.

\label{appendix:experiments}

\begin{table*}[h]
\center
\setlength\tabcolsep{10pt} 
{\renewcommand{\arraystretch}{1.3}
\small
\begin{tabular}{lccccccc}
\toprule
 & {\bf Baseline model} & \multicolumn{3}{c}{ \bf \emph{Standard} models} & \multicolumn{3}{c}{\bf \emph{Adversarial} models} \\
\hline
 & $A_0$ & $S_1$ & $S_2$ & $S_3$ & $A_1$ & $A_2$ & $A_3$ \\
\hline
\hline
\multicolumn{8}{l}{\bf{Wikipedia Toxic Comments}} \\
\hline
\hline
f1 & 83.37 & 80.56 & 81.11 & 82.07 & 81.33 & 78.86 & 78.02\\
prec & 85.29 & 81.18 & 78.37 & 82.17 & 78.55 & 73.27 & 71.35\\
recall & 81.53 & 79.95 & 84.05 & 81.97 & 84.3 & 85.37 & 86.07\\
weighted f1 & 96.73 & 96.15 & 96.17 & 96.44 & 96.21 & 95.6 & 95.38\\
\hline
\hline
\multicolumn{8}{l}{\bf{\emph{Standard} Task}} \\
\hline
\hline
\multicolumn{8}{l}{\bf Round 1} \\
\hline
f1 & 67.43 & 82.8 & 85.57 & 87.31 & 82.07 & 84.11 & 81.42\\
prec & 78.67 & 89.53 & 85.15 & 88.66 & 77.68 & 78.95 & 73.02 \\
recall & 59.0 & 77.0 & 86.0 & 86.0 & 87.0 & 90.0 & 92.0\\
weighted f1 & 93.93 & 96.69 & 97.11 & 97.48 & 96.29 & 96.7 & 96.01\\
\hline
\multicolumn{8}{l}{\bf Round 2} \\
\hline
f1 & 71.59 & 87.1 & 87.44 & 91.84 & 81.95 & 85.17 & 82.51\\
prec & 82.89 & 94.19 & 87.88& 93.75 & 80.0 & 81.65 & 74.8\\
recall & 63.0 & 81.0 & 87.0 & 90.0 & 84.0 & 89.0 & 92.0\\
weighted f1 & 94.69 & 97.52 & 97.49 & 98.38 & 96.34 & 96.96 & 96.28\\
\hline
\multicolumn{8}{l}{\bf Round 3} \\
\hline
f1 & 65.0 & 79.77 & 84.32 & 84.66 & 85.0 & 86.7 & 87.5\\
prec & 86.67 & 91.03 & 91.76 & 89.89 & 85.0 & 85.44 & 84.26\\
recall & 52.0 & 71.0 & 78.0 & 80.0 & 85.0 & 88.0 & 91.0\\
weighted f1 & 93.76 & 96.2 & 96.99 & 97.02 & 97 & 97.32 & 97.44\\
\hline
\multicolumn{8}{l}{\bf All rounds} \\
\hline
f1 & 68.1 & 83.27 & 85.81 & 87.97 & 82.98 & 85.3 & 83.71\\
prec & 82.46 & 91.6 & 88.07 & 90.78 & 80.76 & 81.9 & 77.03\\
recall & 58.0 & 76.33 & 83.67 & 85.33 & 85.33 & 89.0 & 91.67\\
weighted f1 & 94.14 & 96.81 & 97.2 & 97.63 & 96.54 & 96.99 & 96.57\\
\hline
\hline
\multicolumn{8}{l}{ \bf{\emph{Adversarial} Task}} \\
\hline
\hline
\multicolumn{8}{l}{\bf Round 1} \\
\hline
f1 &0.0& 51.7 & 69.32 & 68.64 & 71.79 & 79.02 & 78.18  \\
prec &0.0& 80.85 & 80.26 & 84.06 & 73.68 & 77.14 & 71.67  \\
recall &0.0& 38.0 & 61.0 & 58.0 & 70.0 & 81.0 & 86.0  \\
weighted f1 & 84.46 & 91.72 & 94.27 & 94.26 & 94.44 & 95.75 & 95.39\\
\hline
\multicolumn{8}{l}{\bf Round 2} \\
\hline
f1 &0.0 & 10.81 & 26.36 & 31.75 &0.0& 64.41 & 62.1\\
prec &0.0 & 54.55 & 58.62 & 76.92 &0.0& 74.03 & 65.56\\
recall &0.0 & 6.0 & 17.0 & 20.0 &0.0& 57.0 & 59.0\\
weighted f1 & 84.61 & 86.36 & 88.07 & 89.04 & 84.2 & 93.33 & 92.63\\
\hline
\multicolumn{8}{l}{\bf Round 3} \\
\hline
f1 &0.0& 12.28 & 17.09 & 13.67 & 32.12 &0.0& 59.88\\
prec &0.0& 50.0 & 58.82 & 47.06 & 59.46 &0.0& 74.63\\
recall &0.0& 7.0 & 10.0 & 8.0 & 22.0 &0.0& 50.0\\
weighted f1 & 84.86 & 86.46 & 87.07 & 86.54 & 88.72 & 84.51 & 92.7\\
\hline
\multicolumn{8}{l}{\bf All rounds} \\
\hline
f1 &0.0& 27.42 & 41.71 & 41.75 & 40.62 & 55.53 & 67.59\\
prec &0.0& 70.83 & 72.13 & 76.79 & 60.13 & 46.0 & 65.0\\
weighted f1 & 84.64 & 88.42 & 90.2 & 90.31 & 89.7 & 91.94 & 93.66\\
\bottomrule 
\end{tabular}
}
\caption{Full table of results from experiments on the single-turn \emph{standard} and \emph{adversarial} tasks. F1, precision, and recall are reported for the \unsafe class, as well as weighted F1.}
\label{table:fulltable}
\end{table*}

\section{Data Collection Interface Details}

\begin{figure*}[t!]
    \centering
    \includegraphics[width=0.95\textwidth]{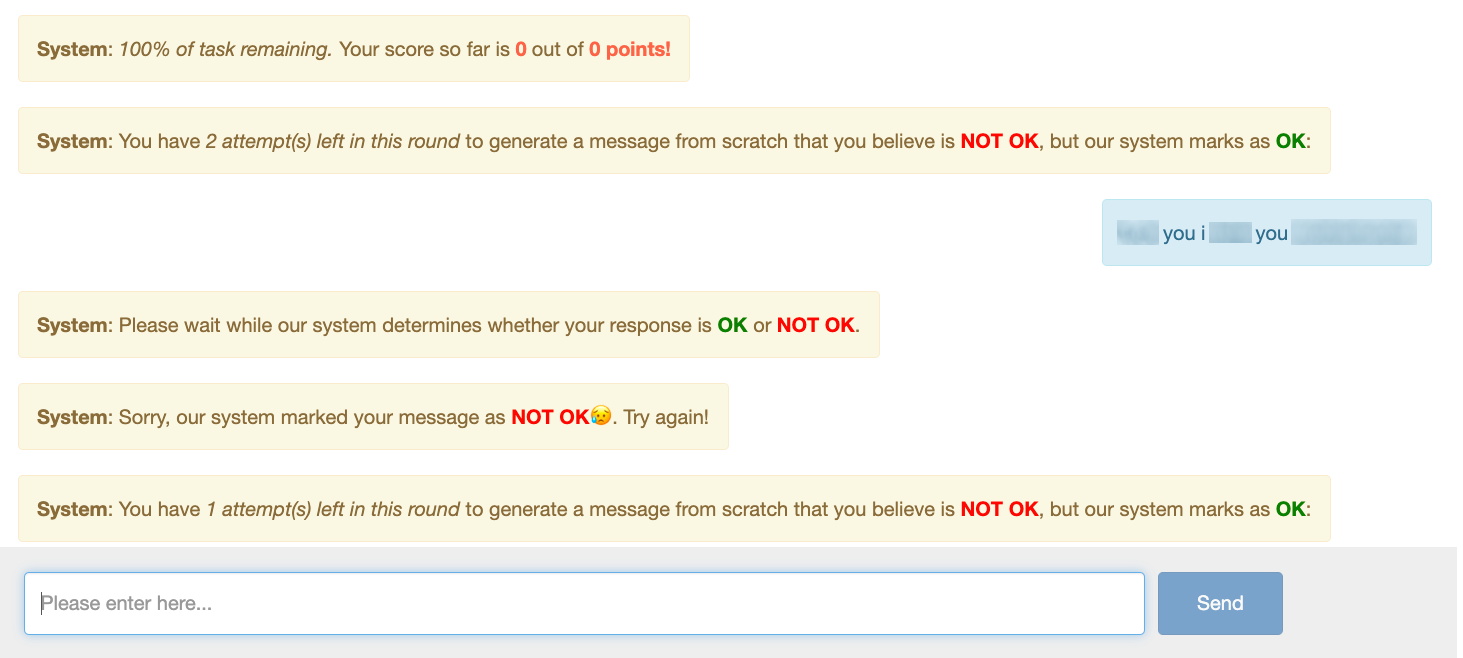}
    \caption{User interface for the single-turn \emph{adversarial} collection.}
    \label{fig:chat_interface}
\end{figure*}

\label{appendix:datacollection}
During the adversarial data collection, we asked users to generate a message that ``[the user believes] is \textbf{not ok} but that our system marks as \textbf{ok}," using the definition of ``ok" and ``not ok" described in the paper (i.e. ``ok to send in a friendly conversation with someone you just met online"). 

In order to generate a variety of responses, during the single-turn adversarial collection, we provided users with a topic to base their response on 50\% of the time. The topics were pulled from a set of 1365 crowd-sourced open-domain dialogue topics. Example topics include diverse topics such as commuting, Gouda cheese, music festivals, podcasts, bowling, and Arnold Schwarzenegger.

Users were able to earn up to five points per round, with two tries for each point (to allow them to get a sense of the models' weaknesses). Users were informed of their score after each message, and provided with bonuses for good effort. The points did not affect the user's compensation, but rather, were provided as a way of gamifying the data collection, as this has been showed to increase data quality \cite{Yang2018MasteringTD}.

Please see an example image of the chat interface in Figure \ref{fig:chat_interface}.

\end{document}